\documentclass{article}

\usepackage{arxiv}

\usepackage[utf8]{inputenc} 
\usepackage[T1]{fontenc}    
\usepackage{hyperref}       
\usepackage{url}            
\usepackage{booktabs}       
\usepackage{amsfonts}       
\usepackage{nicefrac}       
\usepackage{microtype}      
\usepackage{lipsum}
\usepackage{graphicx}
\graphicspath{ {./images/} }

\usepackage{color}
\usepackage{soul}
\usepackage{algpseudocode}
\usepackage{comment}
\usepackage{wrapfig}
\usepackage{algorithm}
\usepackage{amsmath,amssymb,amsthm}

\usepackage{multirow}
\usepackage{makecell}
\usepackage{mathrsfs}
\usepackage{varioref}
\usepackage{cleveref}
\usepackage{xspace}
\usepackage{bm}
\usepackage{enumitem}
\usepackage{pifont}
\usepackage{mathtools}
\usepackage{natbib}
\usepackage{tabularx}
\usepackage{geometry}
\usepackage{array}

\usepackage{subcaption}
\usepackage[table,xcdraw]{xcolor} 

\definecolor{headerbg}{RGB}{44, 62, 80}
\definecolor{thinkingbg}{RGB}{235, 245, 251}
\definecolor{thinkingtx}{RGB}{40, 116, 166}
\definecolor{nothinkingbg}{RGB}{233, 247, 239}
\definecolor{nothinkingtx}{RGB}{30, 132, 73}

\newcommand{\hlthink}[1]{\colorbox{thinkingbg}{\textcolor{thinkingtx}{\ttfamily\bfseries #1}}}
\newcommand{\hlnothink}[1]{\colorbox{nothinkingbg}{\textcolor{nothinkingtx}{\ttfamily\bfseries #1}}}

\title{Thinking vs. NoThinking: Towards Interpreting Reasoning Mechanisms of Large Language Models via Sparse Autoencoders}

\author{
 Bo Cheng \\
  School of Artificial Intelligence, Jilin University\\
  \texttt{chengbo9691@gmail.com} \\
   \And
 Qiaolin Lu \\
  The Hong Kong Polytechnic University\\
\texttt{qiaolin.lu@connect.polyu.hk}\\
   \And
 Yi Chang \\
  School of Artificial Intelligence, Jilin University\\
  Engineering Research Center of Knowledge-Driven Human-Machine Intelligence, MOE, China\\
  International Center of Future Science, Jilin University\\
\texttt{yichang@jlu.edu.cn}\\
  \And
 Yuan Wu\textsuperscript{\thanks{ Corresponding author}} \\
  School of Artificial Intelligence, Jilin University\\
  \texttt{yuanwu@jlu.edu.cn} \\
}

\begin{document}
\maketitle

\begin{abstract}
While Large Language Models (LLMs) employing Chain-of-Thought (CoT) exhibit superior reasoning capabilities, the neural mechanisms distinguishing this explicit \textit{Thinking} mode from direct answer generation (\textit{NoThinking} mode) remain poorly understood. To deconstruct this cognitive process, we apply Top-K Sparse Autoencoders (SAEs) to the intermediate representations of DeepSeek-R1-Distill-Qwen-7B and examine the model's divergent behaviors across math-solving tasks of three distinct difficulty levels. 
Observationally, we identify a clear distinction in how the model functions under two reasoning modes: \textit{Thinking} mode relies on sparse and high-intensity feature activations driving verbal deduction independent of problem complexity, whereas \textit{NoThinking} mode exhibits an adaptive and diffuse pattern prioritizing symbolic manipulation.
Causally, suppressing the three most active sparse features by Total Activation Volume reveals three principles: (i) reasoning and syntactic structure are tightly coupled, as interventions consistently degrade \LaTeX{} and boxed-solution formatting; (ii) \textit{Thinking} responds to disruption with compensatory over-generation marked by increased metacognitive cues and repetitive, low-information continuations; and (iii) coherent CoT behavior depends on a fragile coordination among specialized features, yielding distinct failure modes under perturbation but a consistently impaired output structure.
\end{abstract}

\section{Introduction}

Recent advancements in Large Language Models (LLMs) have demonstrated that explicitly eliciting a Chain-of-Thought (CoT) significantly enhances performance on complex reasoning tasks. Models such as DeepSeek-R1 \citep{2025deepseek} exemplify this paradigm by generating extended thinking process involving exploration, backtracking, and self-correction before producing a final answer. While the behavioral benefits of CoT are well-documented, the underlying neural mechanisms remain opaque \citep{wei2022chain, turpin2023language, lightman2023let}. Specifically, the structural distinctions between the internal states governing explicit reasoning (\textit{Thinking} mode) and direct answer generation (\textit{NoThinking} mode) have yet to be fully elucidated \citep{chen2025does, theodorus2025finding, nanda2023progress,li2022emergent, zou2023representation}. A critical and unresolved issue is whether this thinking process constitutes a distinct computational regime or functions merely as a prolonged extension of standard sequence generation.

To deconstruct this black box, Sparse Autoencoders (SAEs) have emerged as a powerful microscopic tool \cite{rajamanoharan2024improving,cunningham2023sparse}. Pioneering works have successfully addressed the polysemanticity of dense activations by decomposing them into interpretable and monosemantic features \citep{2023relu, cunningham2023sparse, li2025feature, meng2022locating}. With recent architectural innovations such as the Top-K activation mechanism \citep{gao2024scaling} and JumpReLU \citep{lieberum2024gemma}, SAEs have been scaled to analyze massive open-weights models such as Gemma 2. While existing research has made significant strides in dictionary learning, safety auditing \citep{gallifant2025sparse}, and model steering \citep{arad2025saes}, these studies focus predominantly on identifying static semantic concepts or manipulating output logits. Few studies have utilized SAEs to dynamically decode the temporal evolution of reasoning processes or causally disentangle the neural circuits that govern the initiation and regulation of CoT.

Bridging this gap, we apply Top-K SAEs to analyze the intermediate representations of DeepSeek-R1-Distill-Qwen-7B. Unlike prior work that analyzes features in isolation, we establish a comparative framework to contrast the feature dynamics between \textit{Thinking} and \textit{NoThinking} modes processing identical mathematical problems.

Our investigation into the latent feature space reveals a fundamental mechanistic divergence between the two modes. We observe that \textit{Thinking} mode operates through a sparse yet high-intensity activation regime, where a dedicated subset of features drives verbal deduction. Crucially, this reasoning pathway remains stable and invariant to problem complexity. In contrast, \textit{NoThinking} mode exhibits a diffuse and adaptive pattern, recruiting a broader and variable coalition of features to prioritize symbolic manipulation. This strategy bypasses explicit reasoning in favor of difficulty-dependent pattern matching and syntactic retrieval.

Causally, to test whether the discovered sparse features are functionally necessary for \textit{Thinking}, we perform targeted suppression on the top-3 features ranked by Total Activation Volume (TAV). These interventions reveal three governing principles. 
(i) Coupling between reasoning and syntactic structure. Suppressing high-impact features consistently degrades the model's ability to produce formal mathematical outputs, suggesting that logical computation and structural realization are supported by overlapping representations rather than separable reasoning and formatting modules. 
(ii) Compensatory sequence extension under disruption. When the core feature 28634 is suppressed, the model tends to avoid termination and instead expands the generation with increased metacognitive cueing, producing longer but less informative and more repetitive continuations. 
(iii) Fragile coordination under feature suppression. Suppressing different components induces opposite-signed shifts in monitoring-related signals, yet structural degradation remains consistent, implying that \textit{Thinking} depends on a finely balanced coordination among a small set of specialized, high-intensity features with limited redundancy and is prone to distinct failure modes such as uncontrolled verbosity.
\section{Related Work}

\paragraph{SAE Architecture and Scaling.}
Sparse Autoencoders address the polysemanticity of LLM activations by decomposing dense internal states into sparse, interpretable feature combinations. While early work by \cite{cunningham2023sparse} successfully demonstrated this capability in small language models, scaling was historically hindered by training instability and the prevalence of dead latents. To overcome these challenges, recent research has shifted from $L_1$ regularization to direct sparsity enforcement. \cite{gao2024scaling} introduced the Top-$k$ activation mechanism, which simplifies hyperparameter tuning and significantly reduces dead latents by retaining only the $k$ highest-magnitude features. Building on this, \cite{lieberum2024gemma} proposed JumpReLU to dynamically threshold low-magnitude noise. These innovations have enabled the training of SAEs on state-of-the-art open-weight models, such as the massive Gemma Scope suite spanning up to 27B parameters, establishing robust scaling laws for reconstruction fidelity.

\paragraph{Semantic Validation and Interpretability.}
With robust architectures established, focus has shifted to validating the semantic alignment of learned features. A primary method involves automated interpretability, where strong LLMs generate natural language explanations for features based on maximally activating contexts \cite{gallifant2025sparse, gao2024scaling}. Beyond general semantics, \cite{jing2025lingualens} introduced the LinguaLens framework to rigorously analyze linguistic mechanisms. By leveraging "minimal pair" counterfactuals, they demonstrated that SAE features align with specific theoretical categories across morphology and syntax, confirming that LLMs encode precise linguistic attributes in distinguishable sparse directions.

\paragraph{Mechanistic Analysis and Downstream Utility.}
Recent studies have further utilized SAEs to probe model behavior and enhance downstream tasks through causal intervention. In the context of model steering, \cite{arad2025saes} distinguished between input features (pattern detection) and output features (generation influence), showing that steering is most effective when targeting features with high causal scores on output logits. Regarding application, \cite{gallifant2025sparse} found that binarized SAE features outperform dense states in safety-critical tasks like toxicity detection due to better transferability. Similarly, \cite{park2025decoding} applied SAEs to discretize dense retriever embeddings, enabling Concept-Level Sparse Retrieval (CL-SR) that combines semantic expressiveness with the efficiency of sparse representations.
\section{Preliminaries}
In this section, we provide the formal background for sparse autoencoders, with a specific focus on the Top-K sparse autoencoders.

\subsection{Problem Setup}
Let $\mathbf{x} \in \mathbb{R}^{d}$ denote the input vector originating from a specific layer of a pre-trained language model. While typically dense and polysemantic, our aim is to decompose $\mathbf{x}$ into a sparse linear combination of interpretable feature directions from an overcomplete dictionary. Formally, we seek to learn a dictionary matrix $\mathbf{W}_{\text{dec}} \in \mathbb{R}^{d \times m}$ and latent activations $\mathbf{z} \in \mathbb{R}^{m}$ with $m \gg d$ such that the input vector is approximately $\mathbf{x} \approx \mathbf{W}_{\text{dec}} \mathbf{z} + \mathbf{b}_{\text{dec}}$.

\subsection{Sparse Autoencoders}
Sparse Autoencoders implement the decomposition using an encoder-decoder architecture. The encoder maps the input vector $\mathbf{x}$ to latent activations $\mathbf{z}$ via an affine transformation followed by a non-linear activation function $\sigma(\cdot)$:
\begin{equation} 
\mathbf{z} = \sigma(\mathbf{W}_{\text{enc}} \mathbf{x} + \mathbf{b}_{\text{enc}}) 
\end{equation}
where $\mathbf{W}_{\text{enc}} \in \mathbb{R}^{m \times d}$ and $\mathbf{b}_{\text{enc}} \in \mathbb{R}^m$ denote the encoder parameters. ReLU is typically employed as $\sigma(\cdot)$ to enforce non-negativity \citep{2023relu}. Then, the decoder reconstructs the input vector using the learned feature directions:
\begin{equation} 
\hat{\mathbf{x}} = \mathbf{W}_{\text{dec}} \mathbf{z} + \mathbf{b}_{\text{dec}} 
\end{equation}
where $\mathbf{W}_{\text{dec}} \in \mathbb{R}^{d \times m}$ and $\mathbf{b}_{\text{dec}} \in \mathbb{R}^d$ represent the decoder parameters. The model is trained to minimize the following composite objective:
\begin{equation} 
\mathcal{L} = | \mathbf{x} - \hat{\mathbf{x}} |_2^2 + \lambda | \mathbf{z} |_1 
\end{equation}
where $| \mathbf{x} - \hat{\mathbf{x}} |_2^2$ quantifies the reconstruction error, while $| \mathbf{z} |_1$ imposes an $L_1$ penalty weighted by the hyperparameter $\lambda$ to enforce sparsity. However, $L_1$ regularization induces a shrinkage bias where the model suppresses the magnitudes of active feature to minimize the total loss, thereby compromising the fidelity of recovered semantic concepts.

\subsection{Top-K Sparse Autoencoders}
To mitigate the shrinkage bias caused by $L_1$ regularization, we adopt $k$-sparse autoencoders in this work \citep{2013topk}. This architecture enforces sparsity directly through the activation mechanism, rather than a soft penalty in the loss.

Specifically, the model imposes a hard constraint by retaining only the $k$ most significant latents. Given the pre-activation $\mathbf{h} = \mathbf{W}_{\text{enc}} \mathbf{x} + \mathbf{b}_{\text{enc}}$, the latent activations are computed via a TopK operator:
\begin{equation}
\mathbf{z} = \text{TopK}(\mathbf{h})
\end{equation}
where the $i$-th element $\mathbf{z}_i$ retains the value of $\mathbf{h}_i$ if and only if $|\mathbf{h}_i|$ ranks among the top $k$ magnitudes in $\mathbf{h}$. Otherwise, it is set to zero. In addition, ReLU can be applied implicitly or explicitly to ensure positive feature activations. The decoding process remains identical to that of the standard SAEs. Since sparsity is strictly enforced by $k$, the loss function simplifies to the reconstruction loss:
\begin{equation}
\mathcal{L} = | \mathbf{x} - \hat{\mathbf{x}} |_2^2
\end{equation}

This architecture effectively decouples sparsity from activation magnitude, allowing the model to learn precise feature strengths without the downward pressure exerted by regularization penalties.
\section{Experiments}
To investigate the behavior of modern reasoning models when solving mathematical problems across three distinct difficulty levels under two inference modes, we first detail our experimental setup, and then present a comprehensive analysis of the observed patterns.

\subsection{Experiment Setup}

\subsubsection{Model}
We employ DeepSeek-R1-Distill-Qwen-7B \citep{2025deepseek} as our primary subject of analysis. Initialized with Qwen2.5-Math-7B and fine-tuned on the outputs generated by DeepSeek-R1, DeepSeek-R1-Distill-Qwen-7B preserves strong reasoning capabilities while offering a computationally tractable scale for interpretability research.

\subsubsection{Dataset}
To capture the sparse features underlying complex reasoning mechanisms, we employ DeepMath-103K \citep{2025deepmath} as the training corpus for the sparse autoencoders. The large-scale dataset comprises 103,000 mathematical problems, primarily sourced from Math StackExchange \footnote{https://math.stackexchange.com}, and is specifically curated to advance reasoning capabilities.

\subsubsection{Training Details}
\paragraph{Inference Modes.} Modern reasoning architectures, such as R1 and R1-Distill-Qwen, typically segregate internal cognition from final output using specific delimiters (e.g., \texttt{<|beginning\_of\_thinking|>} and \texttt{<|end\_of\_thinking|>}). Based on this structure, we adapt two distinct inference modes as described in \citep{2025nothinking}:
\begin{itemize}
    \item \textit{Thinking}: the mode follows the standard generation trajectory, preserving the full chain of thought within the thinking box before producing the final solution and answer.
    \item \textit{NoThinking}: the mode bypasses the explicit reasoning phase. By constraining the decoding process to keep the thinking box empty, the model is guided to generate only the final solution and answer directly.
\end{itemize}

\paragraph{Activation Extraction.} Training Sparse Autoencoders requires dense activations derived from a target language model. We employ the DeepSeek-R1-Distill-Qwen-7b model on the DeepMath-103K corpus and extract activations from the residual stream of the 13th layer, chosen as a representative intermediate layer\footnote{More details are further provided in  Appendix~\ref{appendix:Hyperparameters}}. The collection process operates in two distinct modes: (1) \textit{Thinking} mode, which processes sequences of length 1,024 to capture CoT reasoning, yielding a total of 108M tokens; and (2) \textit{NoThinking} mode, which excludes reasoning traces, resulting in a total of 65M tokens. 

\paragraph{Hyperparameters.} Our training methodology and hyperparameter settings follow primarily established protocols \citep{gao2024scaling,lieberum2024gemma,wuinterpreting}. Specifically, we train a Top-K Sparse Autoencoder for each mode with $C = 2^{16}$ feature vectors\footnote{More details can be found in Appendix~\ref{appendix:Hyperparameters}}. 
We optimize the model using Adam \citep{adam2014method} with a constant learning rate of $1 \times 10^{-3}$, $\beta_1 = 0.9$, $\beta_2 = 0.999$, and $\epsilon = 6.25 \times 10^{-10}$. To ensure training stability, we implement a dynamic sparsity schedule where the Top-K constraint anneals from $K = 200$ to $K = 20$ during the first 50\% of the initial epoch. Training configurations differ slightly by mode: \textit{Thinking} mode is trained with a batch size of 1,024 for 4 epochs, whereas \textit{NoThinking} mode uses a batch size of 128 for 3 epochs.

\subsubsection{Benchmarks}

To systematically verify the behavioral differences and internal feature dynamics of the reasoning model under varying levels of problem complexity, we categorize our evaluation benchmarks into three difficulty levels: \textit{Easy} (AMC23), \textit{Medium} (AIME 24 \& AIME 25 ), and \textit{Hard} (OlympiadBench). More details of the benchmarks can be found in Appendix~\ref{appendix:Benchmarks}.

\subsubsection{Comparison of Two Reasoning Modes}
To reveal the fundamental differences in the model's internal reasoning mechanisms, we compare feature activation patterns across different difficulty levels.

\paragraph{Feature Selection.}
Rather than selecting features randomly, we target the most dominant components of the model's latent representation. We quantify feature importance by computing  \textit{Total Activation Volume} (TAV) for each feature $i$, defined as the sum of activation magnitudes across a validation corpus comprising $N$ tokens: $\text{TAV}_i = \sum_{n=1}^{N} \mathbf{z}_i^{(n)}$, where $\mathbf{z}_i^{(n)}$ denotes the activation of the $i$-th feature for the $n$-th token. Leveraging the metric, we employ mode-specific SAE to compute the average activation strength across three difficulty levels, identifying the global top-20 features with the highest aggregate activity separately for each mode.

\subsubsection{Causal Intervention}
To move beyond correlational analysis and establish the functional necessity of specific sparse features, we design a causal intervention framework targeting the model's internal reasoning process.

\begin{figure*}[t] 
  \centering
  \includegraphics[width=0.9\textwidth]{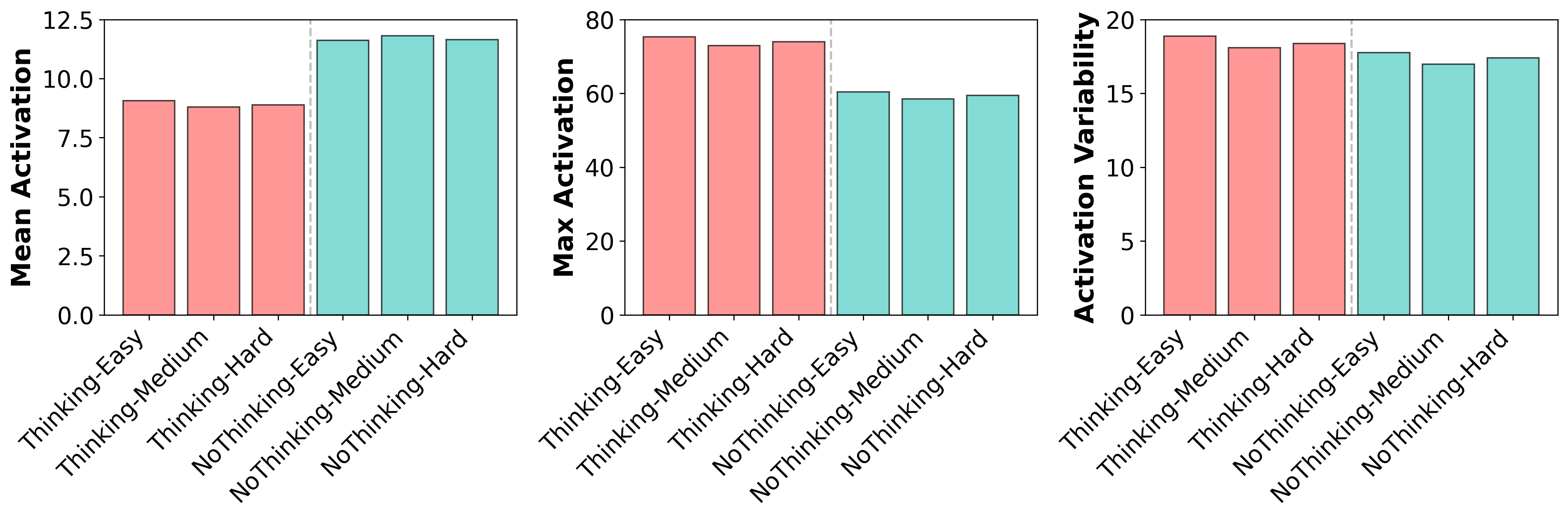}
  \vspace{-0.6em}
  \caption{The mean, maximum, and standard deviation (variability) of activations for the top-20 feature vectors across three difficulty levels. Features are selected based on the TAV metric.}
  \label{fig:exp_feature_activation_stats}
\end{figure*}

\begin{figure*}[htbp]  
  \centering        
  \includegraphics[width=0.9\textwidth]{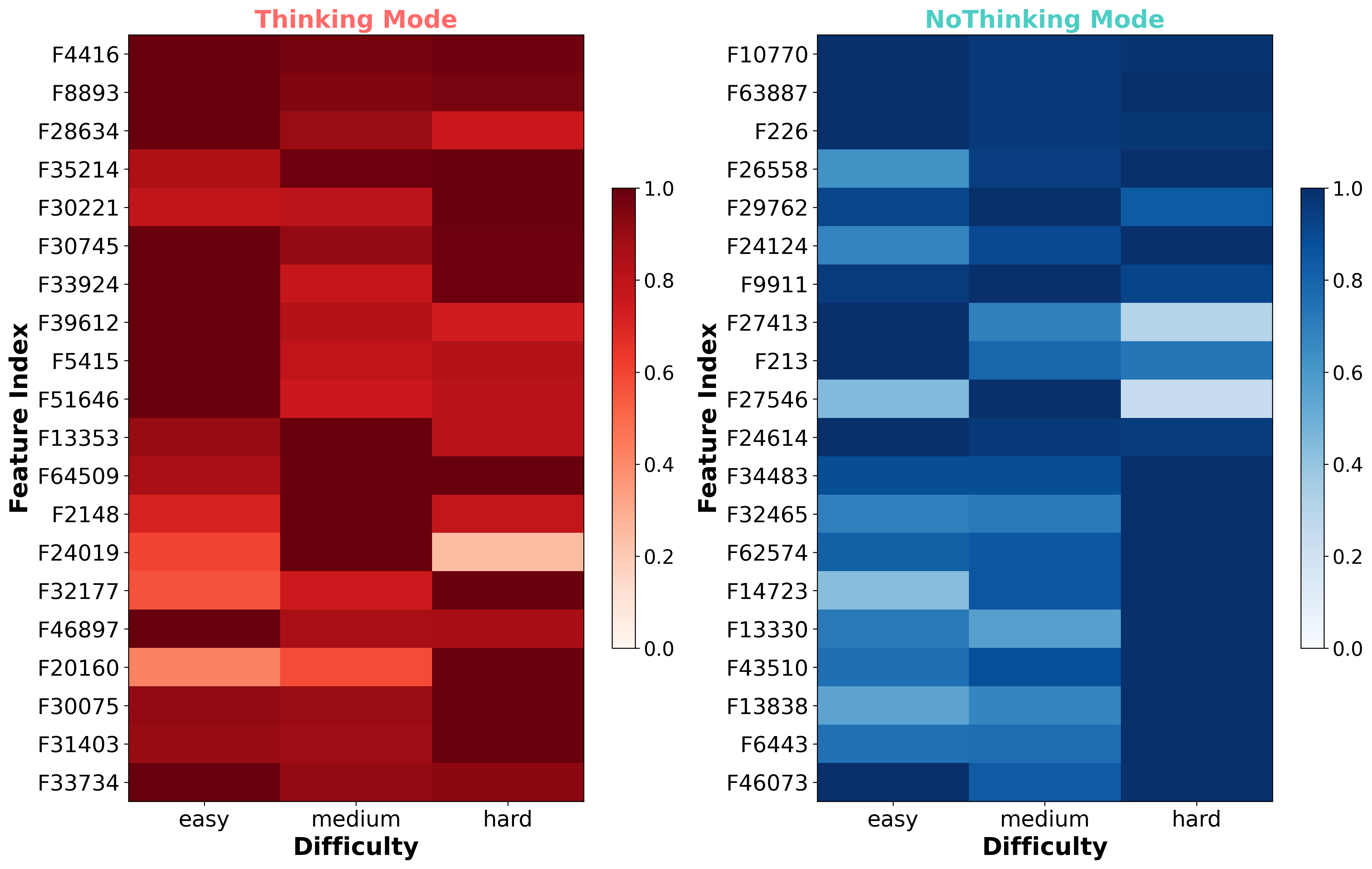}
  \vspace{-0.6em}
  \caption{Heatmaps showing the normalized activations of the top-20 features across difficulty levels for \textit{Thinking} mode and \textit{NoThinking} mode. We apply row-wise normalization, where each feature's activation is scaled by its maximum value across difficulty levels.} 
  \label{fig:exp_feature_activation_heatmap}
\end{figure*}

\paragraph{Feature Selection.}
Guided by the TAV metric and the \textit{Thinking}-specific SAE, we select the top-3 features with the highest aggregate activity on the easy task, specifically Feature 4416, Feature 8893, and Feature 28634, as our primary subjects for intervention. This selection criterion ensures that our analysis targets the neural units that are most salient during the model's reasoning phase.

\paragraph{Intervention Protocol.}
We implement a dynamic suppression hook at the 13th layer, which allows for precise manipulation of feature activations during inference. To isolate the impact on reasoning logic, the intervention is applied exclusively when the model is generating tokens within the thinking block. For a target feature index $i$ and a suppression strength coefficient $\alpha \in [0, 1]$, the modified latent activation $\hat{\mathbf{z}}_{i}^{(t)}$ at time step $t$ is computed as:
\begin{equation}
    \hat{\mathbf{z}}_{i}^{(t)} = (1 - \alpha) \mathbf{z}_{i}^{(t)}
\end{equation}
where $\mathbf{z}_{i}^{(t)}$ represents the original activation value computed by the SAE encoder. We systematically explore four suppression strengths, $\alpha \in \{0.1, 0.3, 0.5, 1.0\}$, ranging from mild attenuation to complete ablation. This graded intervention strategy enables us to characterize non-linear behavioral shifts and identify sensitivity thresholds within the reasoning process.

\paragraph{Evaluation Metrics.}
To quantify the effects of intervention while accounting for the high variance in generated sequence lengths, we define a comprehensive set of density-based metrics capturing linguistic patterns, mathematical formalization and output characteristics, as detailed in Appendix~\ref{appendix:casual_metrics}. All density metrics are normalized per 1,000 tokens to enable fair and robust comparison between baseline and intervened trajectories.

\subsection{Comparison of \textit{Thinking} and \textit{NoThinking}}

\subsubsection{Analysis of Activation Patterns.}

\paragraph{Differences in Activation Patterns.}
The activation statistics presented in Figure~\ref{fig:exp_feature_activation_stats} reveal distinct numerical patterns between the two reasoning modes. Specifically, \textit{Thinking} mode exhibits a relatively low mean activation of approximately 9.0 across all difficulty levels. However, its maximum activation consistently reaches high values around 75.0, accompanied by a high standard deviation of approximately 19.0. This indicates a highly sparse distribution where a small number of features activate intensely and the majority remain suppressed. In contrast,  \textit{NoThinking} mode maintains a significantly higher mean activation of approximately 11.7, with a lower maximum activation stabilizing around 60.0. The slightly lower standard deviation of approximately 17.5 suggests a more uniform and diffuse feature activity. 

\begin{figure*}[t]  
  \centering        
  \includegraphics[width=0.85\textwidth]{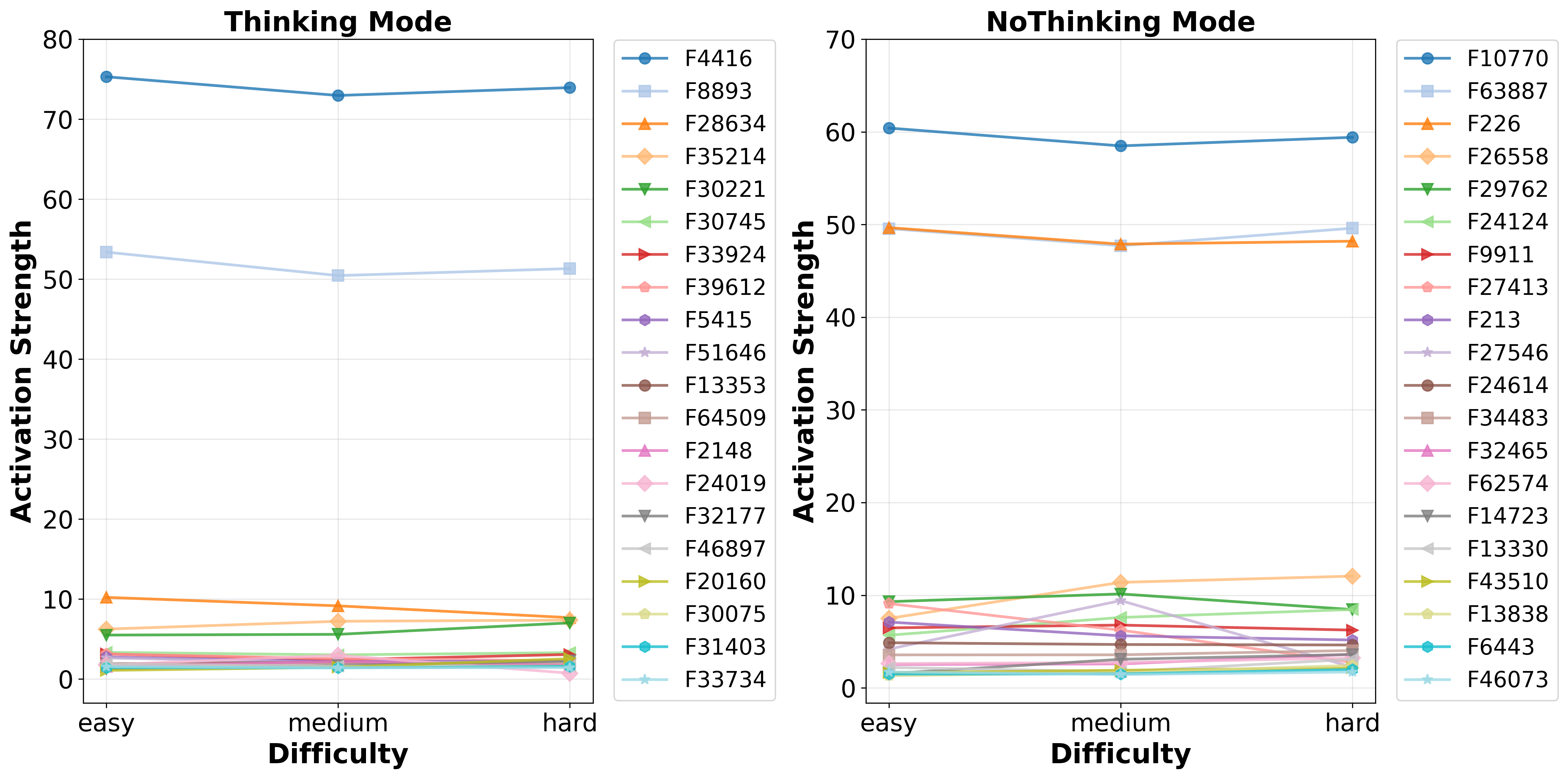}
  \vspace{-0.6em}
  \caption{The activation trajectories of the top-20 SAE features across difficulty levels for both \textit{Thinking} and \textit{NoThinking} modes, highlighting absolute differences in feature intensity.}
  \label{fig:exp_feature_activation_by_difficulty}
\end{figure*}

\paragraph{Consistency of Dominant Feature Intensity.} The activation intensity of dominant features exhibits intrinsic stability regardless of problem complexity, as evidenced by both visual patterns and quantitative metrics in Figures~\ref{fig:exp_feature_activation_heatmap} and~\ref{fig:exp_feature_activation_by_difficulty}. Visually, the continuous horizontal bands in Figure~\ref{fig:exp_feature_activation_heatmap} confirm that the features dominating easy tasks retain their prominence in hard scenarios. Quantitatively, Figure~\ref{fig:exp_feature_activation_by_difficulty} substantiates this robustness. In \textit{Thinking} mode, the dominant F4416 maintains high-intensity activation with values of 75.2 on easy tasks, 72.8 on medium tasks, and 73.9 on hard tasks, significantly outpacing the second-ranked F8893 ($\approx$53.0). This stability in peak magnitude extends to \textit{NoThinking} mode, where the primary F10770 shifts minimally from 60.5 on easy tasks to 59.5 on hard tasks. Overall, while the broader feature composition may evolve, increased difficulty does not induce significant shifts in the activation magnitude of the primary features.

\begin{table*}[htbp]
    \centering
    \small
    \begin{tabular}{@{}ll>{\centering\arraybackslash}p{10cm}@{}} 
        \toprule
        \textbf{Mode} & \textbf{Difficulty} & \textbf{Feature Composition} \\ 
        \midrule
        \multirow{3}{*}{\textit{Thinking}} & \textit{Easy} & \textbf{F4416 (82.0\%)}, F8893 (18.0\%) \\
         & \textit{Medium} & \textbf{F4416 (96.0\%)}, F8893 (4.0\%) \\
         & \textit{Hard} & \textbf{F4416 (100.0\%)}, F8893 (0.0\%) \\ 
        \cmidrule(l){2-3} 
        \multirow{3}{*}{\textit{NoThinking}} & \textit{Easy} & \textbf{F10770 (37.0\%)}, F63887 (25.0\%), F226 (20.0\%), F9911 (18.0\%) \\
         & \textit{Medium} & \textbf{F10770 (83.0\%)}, F63887 (15.0\%), F226 (1.0\%), F9911 (1.0\%) \\
         & \textit{Hard} & \textbf{F10770 (100.0\%)}, F63887 (0.0\%), F226 (0.0\%), F9911 (0.0\%) \\ 
         \bottomrule
    \end{tabular}
    \vspace{-0.4em}
    \caption{Feature source distribution for top-100 highest-activation tokens. Tokens are ranked by activation strength across the top-20 SAE features to identify dominant feature contributions for each mode-difficulty pair.}
    \label{tab:feature_distribution_detailed}
\end{table*}

\subsubsection{Token-Level Analysis of SAE Feature Activations}
While feature-level analysis reveals macroscopic differences between reasoning modes, understanding which specific tokens trigger these features provides crucial interpretability insights. This analysis investigates the lexical characteristics of tokens that maximally activate top-20 SAE features, enabling fine-grained comparison between \textit{Thinking} and \textit{NoThinking} modes across problem difficulties.

\paragraph{Feature Source Distribution.} We first analyze the feature source distribution to determine the extent to which the model relies on specialized latent features. We calculate the composition of the top-100 highest-activation tokens for each mode-difficulty pair and summarize the results in Table~\ref{tab:feature_distribution_detailed}.

As shown in Table~\ref{tab:feature_distribution_detailed}, both modes ultimately converge to single-feature dominance at the hard level, but exhibit fundamentally different trajectories. \textit{Thinking} mode demonstrates early concentration. Even at the easy level, 82\% of top tokens originate from a single feature F4416, increasing monotonically to 100\% at the hard level. In contrast, \textit{NoThinking} mode transitions from a distributed state. Tokens at the easy level disperse across four features, with the top feature F10770 comprising only 37\%. This distribution progressively consolidates into single-feature dominance. These findings suggest that \textit{Thinking} mode employs a stable computational pathway regardless of complexity, whereas \textit{NoThinking} mode adaptively recruits different feature combinations based on task demands.

\begin{wraptable}{r}{0.48\textwidth}
\vspace{-10pt} 
    \centering
    \small
    \renewcommand{\arraystretch}{1.1}
    \setlength{\tabcolsep}{2.2pt}

    \resizebox{\linewidth}{!}{%
    \begin{tabular}{@{}lcccccc@{}}
    \toprule
     & \multicolumn{3}{c}{\textit{\textbf{Thinking}}}
     & \multicolumn{3}{c}{\textit{\textbf{NoThinking}}} \\
    \cmidrule(lr){2-4}
    \cmidrule(l){5-7}

    \textbf{Category}
    & \textit{Easy}
    & \textit{Medium}
    & \textit{Hard}
    & \textit{Easy}
    & \textit{Medium}
    & \textit{Hard} \\
    \midrule

    Word        & 28.2 & 29.4 & 37.7 & 21.7 & 32.4 & 36.1 \\
    Number      & 17.0 & 20.5 & 8.0  & 16.0 & 12.8 & 5.9  \\
    Math Symbol & 15.6 & 14.6 & 13.1 & 22.3 & 14.5 & 14.0 \\
    Reasoning   & 6.1  & 5.5  & 6.9  & 2.9  & 5.3  & 4.4  \\
    Variable    & 7.1  & 7.7  & 10.8 & 6.6  & 6.8  & 9.7  \\
    \bottomrule
    \end{tabular}%
    }

    \caption{Token category distribution comparison across difficulty levels in two reasoning modes (\%), with all tokens activating top-20 SAE features involved.}
    \label{tab:token_category_distribution}
    \vspace{-15pt} 
\end{wraptable}

\paragraph{Token Category Distribution.} We further categorize all activated tokens into functional groups to elucidate their semantic roles, with statistics detailed in Table \ref{tab:token_category_distribution}. The complete categorization taxonomy is provided in Appendix~\ref{appendix:token_taxonomy}. Distinct patterns regarding strategic preference and difficulty adaptation are observed from these statistics. 

First, the modes exhibit a fundamental divergence between verbal deduction and symbolic manipulation. \textit{Thinking} mode relies on explicit verbalized logic by activating a high proportion of reasoning tokens including logical connectives (e.g., "therefore", "implies") and procedural markers. In easy tasks, the proportion of these tokens in \textit{Thinking} mode is nearly double that of \textit{NoThinking} mode. In contrast, \textit{NoThinking} mode employs a formula-centric strategy characterized by a heavy reliance on math symbols. Its usage rate of 22.3\% in simple tasks significantly exceeds the 15.6\% observed in \textit{Thinking} mode.

Second, both modes demonstrate a consistent transition from numerical processing to conceptual abstraction as task difficulty increases. In \textit{Thinking} mode, the frequency of number tokens declines from 17.0\% to 8.0\%, while word tokens increase from 28.2\% to 37.7\%. \textit{NoThinking} mode mirrors this pattern with number tokens falling from 16.0\% to 5.9\% and word tokens rising from 21.7\% to 36.1\%. These results indicate that high-difficulty tasks necessitate linguistic reasoning rather than direct numerical computation regardless of the specific generation strategy.

\paragraph{Qualitative Context Analysis.} By examining specific activation contexts, we identify distinct functional roles for the dominant features, as detailed in Table \ref{tab:Qualitative_Context_Analysis}. F4416 serves as a semantic proxy for \textit{exploratory reasoning} and \textit{self-correction}. It exhibits strong activation during intermediate computational steps (e.g., \textit{"8.97 squared is..."}) and aligns with epistemic markers such as \textit{"Let's see"} or \textit{"isn't working"}, effectively capturing the iterative and trial-and-error nature of the cognitive process. F10770 functions primarily as a \textit{syntactic formatter}. Its activations are densely concentrated on LaTeX syntax (e.g., \texttt{\textbackslash infty}) and formal notation. This indicates that \textit{NoThinking} mode bypasses intermediate logical derivation, prioritizing the retrieval and structural formatting of the final solution.


\begin{table*}[t]
    \centering
    \renewcommand{\arraystretch}{1.15} 
    \small 
    \setlength{\tabcolsep}{5pt} 

    \begin{tabular}{p{0.49\textwidth} p{0.49\textwidth}} 
        \toprule
        \rowcolor{headerbg}
        \multicolumn{1}{c}{\textcolor{white}{\sffamily\bfseries \textit{Thinking} Mode (F4416)}} &
        \multicolumn{1}{c}{\textcolor{white}{\sffamily\bfseries \textit{NoThinking} Mode (F10770)}} \\

        \rowcolor{gray!10}
        \textbf{\footnotesize Functional Role} & \textbf{\footnotesize Functional Role} \\
        \textit{Reasoning Monitor}: Focuses on dynamic and trial-and-error solution finding processes. &
        \textit{Structural Formatter}: Focuses on syntax retrieval and structural formatting of the final solution. \\
        
        \addlinespace[0.5em] 

        \rowcolor{gray!10}  
        \textbf{\footnotesize Representative Activations} & \textbf{\footnotesize Representative Activations} \\

        \begin{itemize}[leftmargin=*, nosep, labelsep=2pt, itemsep=4pt] 
            \item \textbf{Calculation Trace}: 
            \newline 
            \textit{"...8.97 squared is about 80.4609, still less than \hlthink{8}0.92. 8.98 squared is approximately 80.640..."}
            
            \item \textbf{Epistemic Marker}: 
            \newline 
            \textit{"...Because other pairs might have the last letter as L or something else. Wait, let\hlthink{'s} see..."}
            
            \item \textbf{Error Detection}: 
            \newline 
            \textit{"...Wait, now I'm really confused. Maybe this approach isn\hlthink{'t} working. Let's try to write down the equations..."}
        \end{itemize}
        &
        \begin{itemize}[leftmargin=*, nosep, labelsep=2pt, itemsep=6pt]
            
            \item \textbf{LaTeX Syntax} 
            \newline
            $\dots p_9 = \sum_{m=1}^{\text{\texttt{\textbackslash in\hlnothink{fty}}}} \frac{1}{2^m}\dots$
            \newline
            
            \item \textbf{Structure Block} 
            \newline
            $\dots(306+289)=919 \; \text{\texttt{\textbackslash end\{\hlnothink{align}*\}}}\dots$
            \newline
            
            \item \textbf{Formal Logic} 
            \newline
            $\dots\sin(5x) = \frac{\pi}{2} + \pi m \text{\texttt{\textbackslash impl\hlnothink{ies}}} \;\sin(5x)=\frac{1+2m}{14}\dots$
            
        \end{itemize} \\
        \bottomrule
    \end{tabular}
    \vspace{-0.4em}
    \caption{
    \textbf{Analysis of activation contexts for dominant features F4416 (\textit{Thinking}) and F10770 (\textit{NoThinking}).} Examples illustrate typical activation scenarios, where highlighted substrings denote the tokens with the highest activation scores within the context window.
}
    \label{tab:Qualitative_Context_Analysis}
\end{table*}

\subsection{Causal Analysis}

To establish the functional necessity of the identified sparse features, we conducted causal interventions on the top-3 features with the highest Total Activation Volume (TAV): \textbf{F28634}, \textbf{F4416}, and \textbf{F8893}. As shown in Table~\ref{tab:causal_effects}, our results uncover three fundamental mechanisms governing the model's reasoning process.

\paragraph{Coupling of Reasoning and Syntactic Structure.}
Our experiments indicate a functional link between the explicit reasoning process and the generation of mathematical syntax. As shown in Table~\ref{tab:causal_effects}, suppressing critical features for reasoning severely impairs the model's ability to produce formal output regardless of their specific roles. Specifically, we observe a consistent drop in \LaTeX{} density of $-29.54$ to $-40.28$ per 1,000 tokens alongside a near-total failure to format solutions where Boxed Answer Retention frequently fell to $0\%$. These findings suggest that the sparse features driving the Chain-of-Thought (e.g., F4416, F28634) simultaneously encode the structural representations required for formal output, indicating that reasoning and formatting are not processed by independent modules.

\paragraph{Compensatory Sequence Extension.}
\textit{Thinking} mode exhibits a compensatory mechanism when disrupted. When the core reasoning feature F28634 is suppressed, the model does not terminate generation but instead exhibits an expansion of the output sequence. This is evidenced by a 454\% increase in output length. Notably, this expansion is inversely correlated with generation quality where lexical diversity (Distinct-1) declines by 63\%, indicating that the model produces repetitive and low-information sequences. Furthermore, suppressing F28634 leads to a significant increase in metacognitive density (+34.17). This suggests that when the primary reasoning vector is blocked, the model generates additional epistemic markers (e.g., ``Wait,'' ``Let me think'') and extends the sequence length, attempting to maintain the generative state despite the absence of effective computational progress.

\begin{wraptable}{r}{0.48\textwidth}
\vspace{-10pt}
    \centering
    \small
    \renewcommand{\arraystretch}{0.95}
    \setlength{\tabcolsep}{2.5pt}

    \resizebox{\linewidth}{!}{%
        \begin{tabular}{lccc}
            \toprule
            \textbf{Metric (Change per 1k tokens)}
            & \textbf{F28634}
            & \textbf{F4416}
            & \textbf{F8893} \\
            \midrule

            Metacognitive Density ($\Delta$)
            & +34.17 & -19.38 & -4.33 \\
            Uncertainty Density ($\Delta$)
            & +9.25 & -4.16 & -0.99 \\

            \midrule

            \LaTeX{} Density ($\Delta$)
            & -40.28 & -35.22 & -29.54 \\
            Boxed Answer Retention
            & 0\% & 10\% & 0\% \\

            \midrule

            Output Length Change
            & +454\% & +410\% & +107\% \\
            Lexical Diversity (Distinct-1)
            & -63\% & -37\% & -42\% \\

            \bottomrule
        \end{tabular}%
    }
    \caption{\textbf{Impact of Feature Suppression on Reasoning Metrics.} We report relative changes ($\Delta$) from baseline. Key observations include: (1) a universal decline in mathematical formalism (reduced \LaTeX{} density and boxed answers); (2) divergent shifts in metacognition, which increases for F28634 (+34.17) but decreases for F4416 (-19.38) ; and (3) significant output length expansion (up to +454\%) accompanied by reduced lexical diversity across all groups.}
    \label{tab:causal_effects}
    \vspace{-35pt}
\end{wraptable}

\paragraph{Fragile Coordination under Feature Suppression.}
Finally, our results suggest that \textit{Thinking} mode relies on a fragile coordination among distinct high-impact features rather than a uniformly robust mechanism. Importantly, any significant change in metacognitive density reflects a deviation from a stable state rather than an improvement. For instance, suppressing feature 28634 increases metacognitive density by 34.17 while suppressing feature 4416 decreases the same metric by 19.38. This contrast shows that different features regulate the process in opposite directions. Despite these divergent internal effects, the structural indicators collapsed consistently. We observe that \LaTeX\ density drops sharply and boxed answer retention remains near zero across all features. This pattern indicates a coupled control system where specific features jointly maintain both process monitoring and output structure. Consequently, disrupting any single feature drives the model into distinct failure modes such as uncontrolled verbosity.
\section{Conclusion}
In this work, we use Top-K Sparse Autoencoders to probe intermediate representations in DeepSeek-R1-Distill-Qwen-7B and mechanistically distinguish \textit{Thinking} from \textit{NoThinking}. Observationally, \textit{Thinking} mode relies on sparse and high-intensity feature activations driving verbal deduction independent of problem complexity, whereas \textit{NoThinking} mode exhibits an adaptive and diffuse pattern prioritizing symbolic manipulation. Causally, targeted suppression of the most active sparse features shows that reasoning and output structure are tightly coupled, since interventions consistently disrupt \LaTeX{} and boxed-solution formatting; it also reveals a compensatory response in \textit{Thinking}, where disruption triggers longer and more metacognitively signaled but less informative continuations. Together, these findings characterize Chain-of-Thought as a finely tuned, low-redundancy control regime maintained by coordinated feature interactions rather than a standalone reasoning module, pointing toward feature-level control as a path to more reliable and controllable reasoning behavior.

\bibliography{references}
\bibliographystyle{unsrt} 

\appendix
\section{Experiment Setup}

\subsection{Hyperparameters}
\label{appendix:Hyperparameters}
\paragraph{Selection of Targeted Layer.} We specifically target the residual stream of the 13th layer. This decision follows recent insights from sparse autoencoder research \citep{Anthropic2024scaling, lieberum2024gemma}. These studies identify intermediate layers as the core of high-level reasoning. Early layers mostly handle local syntax. Late layers focus on predicting the next token. In contrast, intermediate layers contain the richest semantic information \citep{meng2022locating}. Therefore, this layer is optimal for capturing the reasoning traces in Chain-of-Thought processes.

\paragraph{Selection of Feature Dimensions $C$.} This choice is governed by the scaling law $C \propto Z^\gamma$ \citep{gao2024scaling}, where $Z$ denotes the number of training tokens, with empirical exponents ranging from $\gamma \approx 0.60$ (GPT-2 Small) to $\gamma \approx 0.65$ (GPT-4). Given the fixed number of feature vectors $C$, the implied scaling exponents are $\gamma \approx 0.62$ for the \textit{NoThinking} mode ($Z \approx 65\text{M}$) and $\gamma \approx 0.60$ for the \textit{Thinking} mode ($Z \approx 108\text{M}$). Both values fall consistently within the established range, validating the rationality of our architectural choices. 

\subsection{Details for Benchmarks}
\label{appendix:Benchmarks}
To systematically verify the behavioral differences and internal feature dynamics of the reasoning model under varying levels of problem complexity, we categorize our evaluation benchmarks into three difficulty levels: \textit{Easy} (AMC23) \footnote{https://huggingface.co/datasets/AI-MO/aimo-validation-amc}, \textit{Medium} (AIME 24 \& AIME 25 )\footnote{https://huggingface.co/datasets/AI-MO/aimo-validation-aime}, and \textit{Hard} (OlympiadBench)\footnote{https://huggingface.co/datasets/Hothan/OlympiadBench}.

\begin{itemize}
    \item \textit{Easy} (AMC23). Comprising 40 problems from the 2023 American Mathematics Competition. It focuses on high-school foundational topics such as algebraic manipulations and geometric principles, with all solutions constrained to integers between 0 and 999.
    
    \item \textit{Medium} (AIME 24 \& AIME 25). Consisting of 30 problems from the 2024 American Invitational Mathematics Examinations and 30 novel problems curated in 2025, respectively. Unlike the AMC, these tasks necessitate deep combinatorial and geometric insights involving multi-step reasoning and significantly higher computational complexity. The answers are strictly constrained to integers from 0 to 999.
    
    \item \textit{Hard} (OlympiadBench). A comprehensive corpus of 8,476 Olympiad-level problems sourced from elite competitions such as the IMO and the Chinese Gaokao, which is much more challenging than AIME and AMC. It is characterized by multi-modal inputs (e.g., diagrams) and expert solutions that involve complex, long-horizon logical chains. We selected a subset of 503 samples from this benchmark for our analysis.
\end{itemize}

\subsection{Token Categorization}
\label{appendix:token_taxonomy}
We define a hierarchical taxonomy of 5 token categories to characterize the semantic properties of highly-activated tokens: (1) reasoning: logical connectives (e.g., \textit{"therefore"}, \textit{"since"}, \textit{"implies"}), procedural markers (e.g., \textit{"first"}, \textit{"step"}, \textit{"then"}), and problem-solving directives (e.g., \textit{"let"}, \textit{"solve"}, \textit{"calculate"}); (2) number: pure digit sequences (e.g., \textit{"123"}, \textit{"2024"}); (3) math symbol: mathematical operators and brackets (e.g., \textit{"+"}, \textit{"="}, \textit{"()"}); (4) variable: single alphabetic characters typically representing mathematical variables (e.g., \textit{"x"}, \textit{"n"}, \textit{"A"}); (5) word: multi-character alphabetic strings (e.g., \textit{"the"}, \textit{"angle"}).

\begin{table*}[t]
    \centering
    \small 
    \renewcommand{\arraystretch}{1.3} 
    
    \begin{tabular}{@{} l l p{9cm} @{}}
        \toprule
        \textbf{Dimension} & \textbf{Metric} & \textbf{Key Indicators \& Description} \\
        \midrule
        
        \multirow{2}{*}{\textbf{Cognitive Ability}} 
        & Metacognitive Density & \textbf{16 Markers:} \textit{"wait", "hmm", "actually", "let's see", "perhaps", "maybe", "alternatively", "hold on", "thinking about", "let me", "i think", "seems like", "looks like", "but wait", "oh wait", "hang on"}. \\
        
        & Uncertainty Density & \textbf{9 Markers:} \textit{"might", "could", "possibly", "probably", "maybe", "perhaps", "seems", "appears", "likely"}. \\
        \midrule
        
        \multirow{2}{*}{\textbf{Math Formalization}} 
        & LaTeX Density & Counts mathematical environments, including both display (\texttt{\textbackslash[...\textbackslash]}) and inline (\texttt{\textbackslash(...\textbackslash)}) syntax usage. \\
        
        & Boxed Answer Retention & A binary indicator measuring the successful generation of the strict final answer format: \texttt{\textbackslash boxed\{...\}}. \\
        \midrule
        
        \multirow{2}{*}{\textbf{Generative State}} 
        & Output Length & Tracks the total token count of the generated chain-of-thought to monitor verbosity changes. \\
        
        & Lexical Diversity & Quantified via \textbf{Distinct-1} \cite{li2016Distinct-1}: the ratio of unique unigrams to total tokens. Low values indicate repetitive looping or mode collapse. \\
        
        \bottomrule
    \end{tabular}
    \caption{\textbf{Summary of Evaluation Metrics for Causal Interventions.} The framework categorizes metrics into cognitive, structural, and generative dimensions. Note that all density metrics are normalized per 1,000 tokens.}
    \label{tab:evaluation_metrics}
\end{table*}

\subsection{Casual Evaluation Metrics}
\label{appendix:casual_metrics}
To quantify the impact of SAE feature interventions on model reasoning behavior, we define a comprehensive set of metrics capturing linguistic patterns, mathematical formalization and output characteristics, as detailed in Table~\ref{tab:evaluation_metrics}. All density metrics are normalized per 1,000 tokens to enable fair comparison across responses of varying lengths. Specifically, we assessed cognitive ability through Metacognitive Density and Uncertainty Density. The former tracks the frequency of 16 specific markers, while the latter considers a set of 9 indicators. In addition, mathematical formalization was evaluated via LaTeX Density, measuring the prevalence of symbolic notation, and Boxed Answer Retention, which serves as a binary indicator of the model's capacity to formulate valid and well-structured conclusions. Furthermore, we analyzed generative characteristics using Output Length Change and Lexical Diversity to detect behavioral anomalies such as repetitive looping or verbose degeneration indicative of reasoning breakdown.

\end{document}